\documentclass[letterpaper, 10 pt, conference]{ieeeconf}  % Comment this line out if you need a4paper

\IEEEoverridecommandlockouts                              % This command is only needed if 
\usepackage{graphicx}

\usepackage{tabularx}
\usepackage{booktabs}
\usepackage{bm}
\usepackage{caption}

\usepackage[table, dvipsnames]{xcolor}

\usepackage{amsmath,amssymb}

\usepackage{hyperref}
\usepackage{flushend}
\usepackage[font=normalsize]{subcaption}

\definecolor{htmlcssgreen}{rgb}{0.0, 0.5, 0.0}
\newcommand{\increase}[1]{\footnotesize \textcolor{htmlcssgreen}{#1}\normalsize}
\newcommand{\grayedout}[1]{{\color{gray}  #1}}
\definecolor{britishracinggreen}{rgb}{0.0, 0.26, 0.15}
\newcommand{\bestresult}[1]{\boldmath{\textcolor{red}{#1}}}
\newcommand{\secondbest}[1]{\textcolor{blue}{\underline{#1}}}

\title{\LARGE \bf
Egocentric RGB+Depth Action Recognition in Industry-Like Settings
}

\author{Jyoti Kini$^{1}$, Sarah Fleischer$^{1}$, Ishan Dave$^{1}$ and Mubarak Shah$^{1}$ % <-this % stops a space
\thanks{$^{1}$Center for Research in Computer Vision, University of Central Florida, USA 
        {\tt\small \{jyoti.kini, sarah.fleischer, ishanrajendrakumar.dave\}@ucf.edu, shah@crcv.ucf.edu}}%
}

\begin{document}

\maketitle
\thispagestyle{empty}
\pagestyle{empty}

%%%%%%%%%%%%%%%%%%%%%%%%%%%%%%%%%%%%%%%%%%%%%%%%%%%%%%%%%%%%%%%%%%%%%%%%%%%%%%%%
\begin{abstract}
% In this work, we propose an ensemble modeling approach for multimodal action recognition. We independently train individual modality models using a variant of focal loss tailored to handle the long-tailed distribution of the MECCANO~\cite{ragusa2021meccano} dataset. Based on the underlying principle of focal loss, which captures the relationship between tail (scarce) classes and their prediction difficulties, we propose an exponentially decaying variant of focal loss for our current task. It initially emphasizes learning from the hard misclassified examples and gradually adapts to the entire range of examples in the dataset. This annealing process encourages the model to strike a balance between focusing on the sparse set of hard samples, while still leveraging the information provided by the easier ones. Additionally, we opt for the {\it{late fusion}} strategy to combine the resultant probability distributions from RGB and Depth modalities for final action prediction. Experimental evaluations on the MECCANO dataset demonstrate the effectiveness of our approach.
Action recognition from an egocentric viewpoint is a crucial perception task in robotics and enables a wide range of human-robot interactions. While most computer vision approaches prioritize the RGB camera, the Depth modality—which can further amplify the subtleties of actions from an egocentric perspective—remains underexplored.
Our work focuses on recognizing actions from egocentric RGB and Depth modalities in an industry-like environment. To study this problem, we consider the recent MECCANO dataset, which provides a wide range of assembling actions.
Our framework is based on the 3D Video SWIN Transformer to encode both RGB and Depth modalities effectively. To address the inherent skewness in real-world multimodal action occurrences, we propose a training strategy using an exponentially decaying variant of the focal loss modulating factor.
Additionally, to leverage the information in both RGB and Depth modalities, we opt for late fusion to combine the predictions from each modality.
We thoroughly evaluate our method on the action recognition task of the MECCANO dataset, and it significantly outperforms the prior work. Notably, our method also secured first place at the multimodal action recognition challenge at ICIAP 2023.

% In order to handle inherent skewness in real-world multimodal action occurrences, we propose a training strategy utilizing an exponentially decaying variant of the modulating factor of the focal loss. 

\end{abstract}

%%%%%%%%%%%%%%%%%%%%%%%%%%%%%%%%%%%%%%%%%%%%%%%%%%%%%%%%%%%%%%%%%%%%%%%%%%%%%%%%
\section{INTRODUCTION}
Recent advancements in action recognition have paved the way for numerous practical applications, ranging from behavioral studies~\cite{behaviour} and sports analytics~\cite{diving, xu2022finediving} to visual security systems~\cite{gabv2, rizve2021gabriella, fioresi2023tedspad}, and systems designed to detect falls in elderly individuals~\cite{buzzelli2020vision, liu2020privacy}. In the realm of robotics, the capability to detect and interpret human actions and gestures is a crucial perception task. This is especially true when robots are expected to engage with humans and execute tasks across various sectors, including manufacturing, healthcare, and service robotics. Actions like pointing, reaching, or grasping become especially pivotal since they often relay valuable insights about the user's requirements and intentions.

While traditional video analysis captures a bulk of human behavior, it occasionally overlooks subtle nuances. This is where egocentric cameras prove beneficial. Offering a unique first-person perspective, these cameras offer a more intimate vantage point of human-object interactions and movements. This detailed view is essential in settings where robots need to closely work with humans and need to understand both their actions and reasons behind them.

\begin{figure}[h]
      
    \begin{subfigure}{0.5\textwidth}
        \vspace{1mm}
        \centering
        \includegraphics[width=0.9\textwidth]{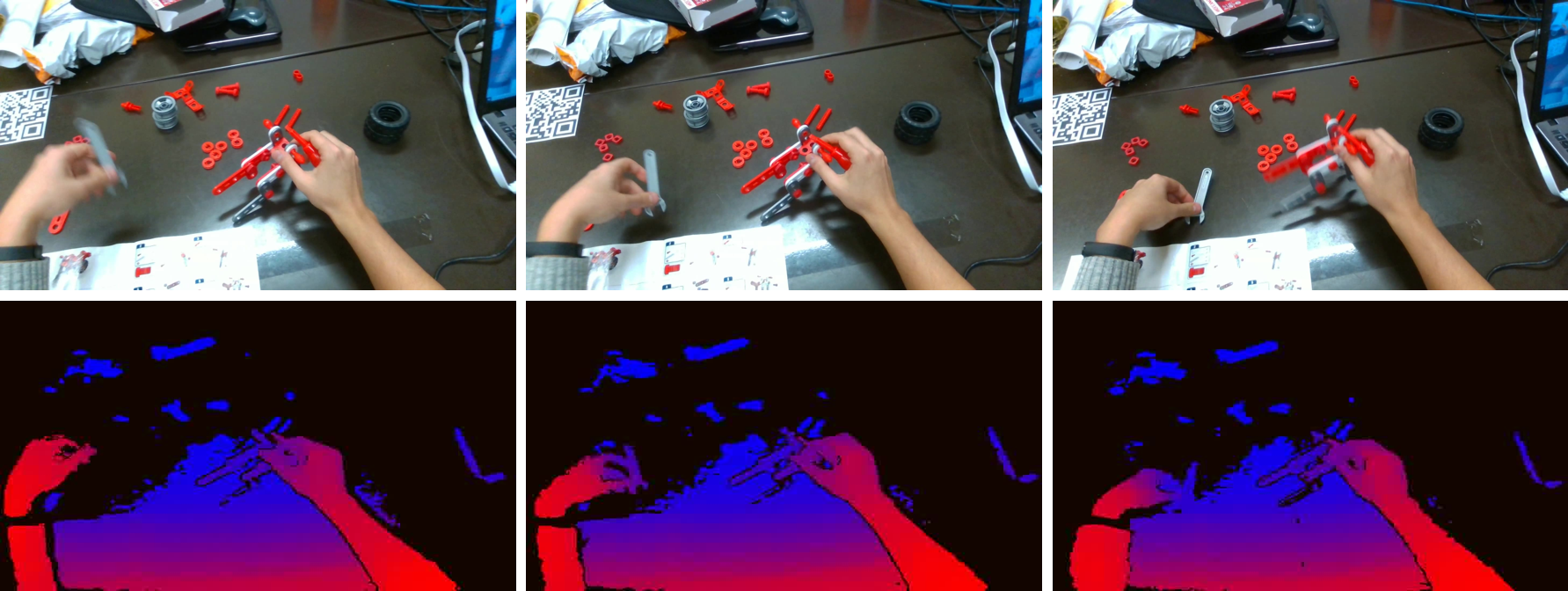}
        % \vspace{-2mm}
        \caption{\texttt{put\_wrench} }

    \end{subfigure}

    \hfill
    \begin{subfigure}{0.5\textwidth}
    \vspace{2mm}
    \centering
        \includegraphics[width=0.9\textwidth]{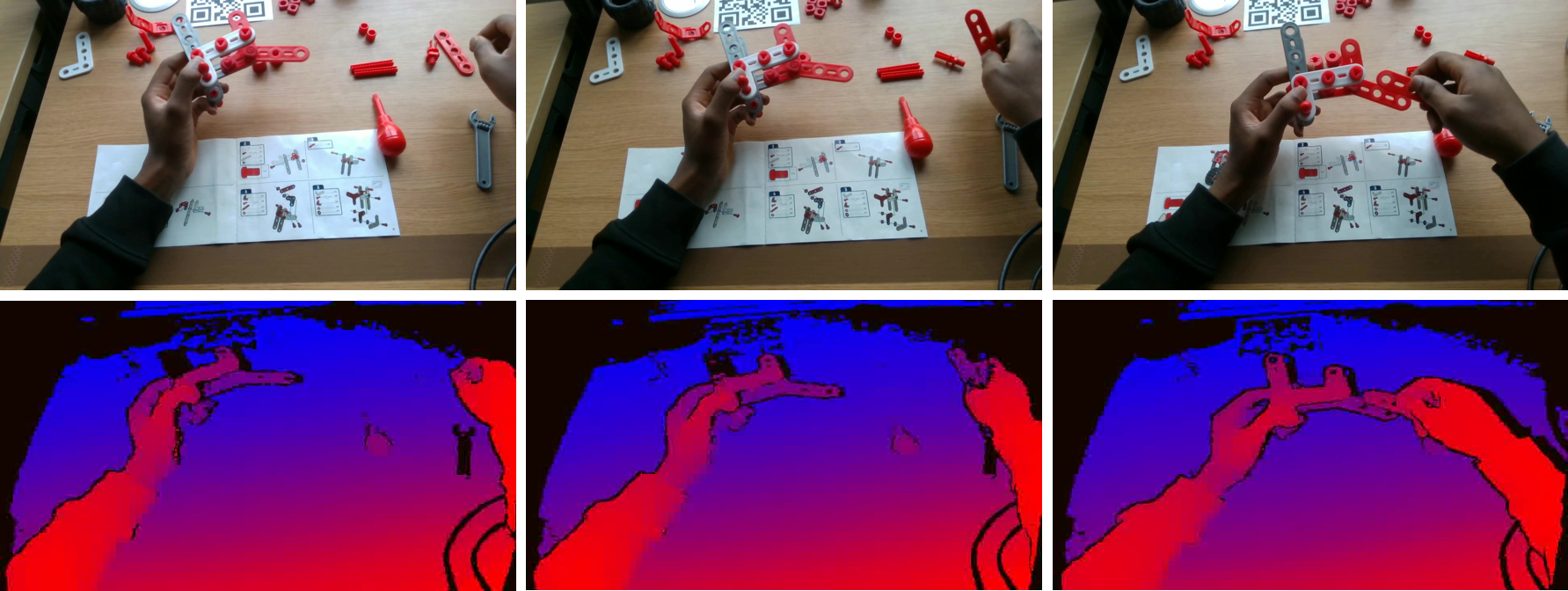}
        % \vspace{-2mm}
        \caption{\texttt{take\_red\_perforated\_bar} }

    \end{subfigure}
    \hfill
    % \vspace{-2mm}
    \caption{\textbf{Sample actions from the MECCANO dataset~\cite{ragusa_MECCANO_2023}.}
    In both examples (a) and (b), the top row shows video frames from RGB modality and the bottom row shows corresponding frames from Depth modality. MECCANO dataset provides various actions from assembling the toy-motorbike in an industry-like setting.}
    \label{fig:visualize}
\end{figure}

A common challenge in egocentric action recognition is the heavy focus on RGB data. Though RGB yields rich visual details, it falls short in conveying depth or spatial relationships. Integrating Depth with RGB fills this gap, providing a more holistic visual understanding. Depth modality offers information about the distance and relation between objects, which is very useful when interpreting actions from an egocentric point of view.

The combination of RGB and Depth data is showcased in the recent MECCANO dataset~\cite{ragusa_MECCANO_2023}, which captures these elements in industry-like settings.
The dataset captures various complex assembly actions of a toy-motorbike, as shown in Fig.~\ref{fig:visualize}. By using this dataset, our research aims to explore how these two modality of data can improve egocentric action recognition.

Prior work, such as UMDR~\cite{zhou2023unifiedpami}, addresses the RGB+Depth action recognition challenge using a video data augmentation strategy. They built upon existing MixUp augmentations to provide temporal regularization. Although this method performs well on third-person datasets, their heavy reliance on augmentation struggles to generalize for real-world egocentric datasets.

For our approach, we first utilize a recent Swin3D \cite{videoswin} video encoder to capture spatio-temporal features from RGB and Depth modalities. We note that real-world multimodal data, associated with action occurrences, exhibit an inherent skewness, leading to a long-tailed action recognition scenario. In such cases, some action classes are prevalent and well-represented in the training data (e.g., \texttt{check\_booklet, align\_screwdriver\_to\_screw, plug\_rod, align\_objects} actions), while others are scarce (e.g., \texttt{fit\_rim\_tire, put\_nut, put\_wheels\_axle} actions), leading to significant data imbalance. The inherent complexity of multimodal data, combined with such data imbalance, presents a formidable challenge for learning approaches.
Based on the underlying principle of focal loss, which captures the relationship between tail (scarce) classes and their prediction difficulties, we propose an exponentially decaying variant of focal loss modulating factor for our current task. It initially emphasizes learning from the hard misclassified examples and gradually adapts to the entire range of examples in the dataset. This annealing process encourages the model to strike a balance between focusing on the sparse set of hard samples, while still leveraging the information provided by the easier ones. Additionally, we opt for the late fusion strategy to combine the resultant probability distributions from RGB and Depth modalities for final action prediction.

Our method is evaluated on the action recognition task of the MECCANO dataset, where it outperforms prior baselines by significant margins and establishes a new benchmark. Our method also secured first place in the multimodal action recognition challenge of \textit{22nd International Conference on Image Analysis and Processing} (ICIAP) 2023.

To summarize, our primary contributions are:
\begin{itemize}
    \item We propose a training framework for multimodal action recognition from an egocentric camera with RGB and Depth modalities.
    \item  We introduce an exponentially decaying variant of focal loss modulating factor to address the challenges of inherent skewness of the multimodal data.
    \item Our method sets the benchmark with state-of-the-art results on the MECCANO dataset and has secured first place in the multimodal action recognition challenge at ICIAP 2023.
\end{itemize}

\begin{figure*}[t]
  \centering
  \vspace{2mm}
  \includegraphics[width=0.95\textwidth]{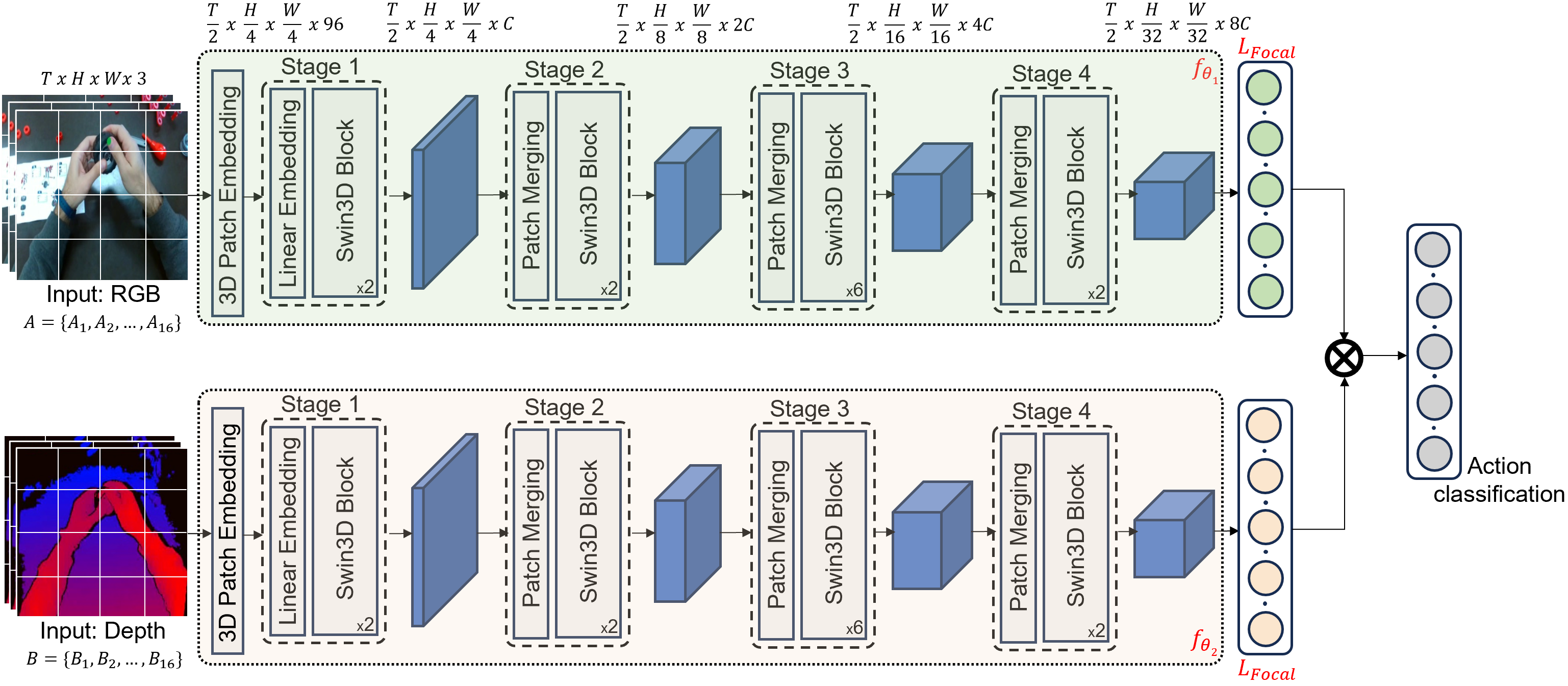}
  \caption{\textbf{Overview of our framework} The RGB frames $\{A_{i}$, $A_{i-1}$,..,$A_{T}\}$ and Depth frames $\{B_{i}$, $B_{i-1}$,..,$B_{T}\}$ are passed through two independently trained Swin3D-B~\cite{videoswin} encoders ${f}_{\theta_1}$ and ${f}_{\theta_2}$ respectively to generate feature tokens. The resultant class probabilities, obtained from each pathway, are averaged to subsequently yield action classes. Exponentially decaying focal loss $L_{Focal}$ is leveraged to deal with the long-tailed distribution exhibited by the data.} 
  \label{fig:Architecture}
  \vspace{-2mm}
\end{figure*} 

\section{Related Work}
% Video Understanding
% Action Recognition
% Ego-Centric Action
% Multi-modal Action
% RGB-D based multimodal action recognition
Video understanding aims to extract meaningful spatio-temporal features from videos. This domain encompasses a wide range of problems such as video object detection, object tracking, action recognition, temporal action localization, action anticipation, repetition counting, and more.
\subsection{Action Recognition}

Developments in action recognition have largely been driven by architectural changes and novel training paradigms. Among the many architectures that have emerged, convolution-based models like C3D~\cite{c3d}, R3D~\cite{kenshohara}, R2plus1D~\cite{r2plus1d}, and X3D~\cite{feichtenhofer2020x3d} set the early benchmarks. Following this, transformer-based models such as TimeSFormer~\cite{gberta_2021_ICML}, VTN~\cite{vtn}, and VideoSWIN~\cite{videoswin} came into the spotlight, offering innovative ways to process video data. Concurrently, the focus on data efficiency led to the exploration of paradigms like self-supervised learning~\cite{stmae, tclr, jenni2021time, Blegiers2023EventTransAct}, semi-supervised learning~\cite{timebalance,semi_cmpl,semi_mvpl}, and few-shot recognition~\cite{samarasinghe2023cdfsl}.

Diverse datasets have significantly propelled the field forward. Examples include Kinetics~\cite{kay2017kinetics}, HVU~\cite{hvu}, HACS~\cite{hacs}, MMiT~\cite{mmit}, MultiSports~\cite{li2021multisports}, UCF101~\cite{soomro2012ucf101}, HMDB51~\cite{hmdb}, and APPROVE~\cite{gupta2023class}. It's worth noting that these datasets predominantly focus on the third-person view.

\subsection{Egocentric Action Recognition}
With the increasing interest in a more intimate and user-centric perspective, several egocentric datasets have been introduced. EPIC-KITCHENS~\cite{Damen_2018_ECCV}, for instance, offers a comprehensive view of daily kitchen activities over multiple days. Charades-Ego~\cite{sigurdsson2018actor} takes a different approach, focusing on the joint modeling of first and third-person videos. The datasets Something-Something~\cite{stst} and 100DOH~\cite{shan2020understanding} delve into the realm of hand-centric activities, offering insights into human hand interactions. HOMAGE~\cite{rai2021home} introduces the use of synchronized multi-view cameras, providing an enriched perspective that includes an egocentric view. Furthermore, Ego4D~\cite{grauman2022ego4d} captures daily-life activities from various global locations, adding a layer of diversity to the data.

\subsection{Multimodal RGB+Depth Based Action Recognition}
While many datasets focus on learning from the visual (RGB) modality, depth cameras bring enhanced spatial recognition, the ability to discern intricate human-object interactions, and the capacity to capture subtleties often missed by RGB cameras. There is a growing body of work exploring action recognition combining RGB and Depth data. Pioneering datasets in this area include NTU RGB-D, a large-scale human interaction dataset~\cite{shahroudy2016ntu}, NvGesture, concentrating on touchless driver controls~\cite{molchanov2016online}, and Chalearn IsoGD, focusing on gesture videos~\cite{wan2016chalearn}. Methods leveraging both RGB and Depth cues, such as those cited in ~\cite{wang2018cooperative, zhou2021regional, wang2017scene,shahroudy2017deep}.
However, most of these resources are not egocentric and don't replicate industrial settings. This observation led us to focus on the MECCANO dataset~\cite{ragusa_MECCANO_2023}, which emulates industry-like actions by showcasing the intricate assembly processes of a toy motorbike. It offers RGB, Depth, and an added Gaze modality for human eye tracking. In our study, we restrict our scope to the RGB and Depth modalities.

UMDR~\cite{zhou2023unifiedpami} emerges as a particularly relevant approach to our RGB+Depth based egocentric action recognition goal and has been submitted to the MECCANO challenge~\cite{meccano_challenge} (Table~\ref{table:Sota}). This method enhances the existing MixUp augmentation for videos by adding motion regularization. Our approach deviates significantly from UMDR. Rather than relying on the video augmentations, we prioritize hard-to-classify tail classes before the head classes, addressing the skewness inherent in multimodal egocentric action occurrences.

\section{Approach}
\subsection{Cross-Modal Fusion}
\label{Cross-Modal Fusion}
Fig.~\ref{fig:Architecture} provides comprehensive details of our proposed approach. Given a set of spatiotemporally aligned RGB and Depth sequences that extend between $[t_s, t_e ]$, where $t_s$ and $t_e$ are the start and the end duration of the sequence, our goal is to predict the action class $\mathcal{O} = \{o_{1}$, $o_{2}$,.., $o_{K}\}$ associated with the sequence. In order to achieve this, we adopt an ensemble architecture comprising two dedicated Video Swin Transformer~\cite{videoswin} backbones to process the RGB clip $\mathcal{A} = \{A_{i}$, $A_{i-1}$,..,$A_{T}\}$ and Depth clip $\mathcal{B} = \{B_{i}$, $B_{i-1}$,..,$B_{T}\}$ independently. Here, $i$ corresponds to a random index spanning between $t_s$ and $t_e$. The input video for each modality defined by size $T\times H\times W\times 3$ results in token embeddings of dimension $\frac{T}{2}\times H_d\times W_d\times 8C$. We pass this representation retrieved from {\it{stage-4}} of the base feature network to our newly added fully connected layer and fine-tune the overall network. The final prediction is derived by averaging the two probability distributions obtained as output from the RGB and Depth pathways. 

\noindent\textit{\underline{Video Swin Transformer}}
We utilize the Video Swin Transformer~\cite{videoswin} backbone to extract spatiotemporal features from input sequences. The input sequence, with dimensions $T\times H\times W\times 3$, is divided into non-overlapping 3D patches/tokens of size $2\times 4\times 4\times 3$. Next, the 3D Patch Embedding layer generates $\frac{T}{2}\times \frac{H}{4}\times \frac{W}{4}$ token with 96-dimensional representation. A Linear Embedding layer is subsequently applied to project these features into a $C$-dimensional space. Thereafter, no subsampling takes place along the temporal dimension. However, in the patch merging layer, features from adjacent patches, organized into $2\times2$ spatial groups, are fused together. This fusion is followed by a linear layer that reduces the dimension of the concatenated features by half. Within the Swin3D Block, a specific attention mechanism known as the 3D shifted window-based multi-head self-attention module is employed. This is followed by a 2-layer MLP, with GELU~\cite{hendrycks2016gaussian} non-linearity in between. The 3D shifted window-based multi-head self-attention module enables attention computation within each 3D window, and facilitates cross-window connections, while efficiently calculating self-attention based on non-overlapping windows. The hierarchical processing of patches and the self-attention mechanisms provided by the Video Swin Transformer enable effective capture of complex spatiotemporal patterns.

\subsection{Focal Loss: Exponentially Decaying Modulating Factor}
\label{Losses}
Focal loss~\cite{lin2017focal} is a variant of cross-entropy loss with a modulating factor that down-weighs the impact of easy examples and focuses on the hard ones. It, therefore, tends to prevent bias towards data-rich classes and improves the performance on scarce categories. \\
Multi-classification cross-entropy (CE) loss is given by:
\begin{align}
    L_{CE} = - \sum_{j=1}^{K}{y_j \log(p_j)} 
\end{align}
where, say we have $K$ action classes, and $y_j$ and $p_j$ correspond to the ground-truth label and predicted probability respectively for the $j^{th}$ class. \\
On the other hand, the key objective of focal loss~\cite{lin2017focal} is defined as:
\begin{align}
    L_{Focal} = - \sum_{j=1}^{K} {(1-p_j)^{\gamma} \log p_j}
\end{align}
In our work, we use focal loss $L_{Focal}$, and exponentially decay the modulating factor $\gamma$ from $\gamma_{init} = 2$ to $\gamma_{fin} = 0.1$ over the entire training duration. \\
The exponential interpolation/decay of $\gamma$ over total epochs $Z$ is defined by:
\begin{align}
    \gamma_{curr} = \gamma_{init} * (\gamma_{fin} / \gamma_{init}) ^ {(z_{curr} / Z)}
\end{align}
where, $\gamma_{curr}$ refers to the current value of the modulating factor at a specific epoch $z_{curr}$. \\
When $\gamma$=0, the objective function is equivalent to cross-entropy loss. Our proposed annealing process for $\gamma$ allows for the model to focus on the sparse set of hard examples in the early stage of training, and gradually shift its focus towards easy examples. This configuration is essential to ensure that the model learns meaningful representations and generalized decision boundaries. \\
% Although unsuccessuWe, also, attempt to use liner interpolation of $\gamma$ given by:
% \begin{align}
%     \gamma_{curr} = \gamma_{init} + (\gamma_{fin} - \gamma_{init}) * {(z_{curr} / Z)}
% \end{align}
% def linear_annealing(current_epoch, total_epochs, initial_gamma, final_gamma):
%     return initial_gamma + (final_gamma - initial_gamma) * (current_epoch / total_epochs)
% def exponential_annealing(current_epoch, total_epochs, initial_gamma, final_gamma):
%     return initial_gamma * (final_gamma / initial_gamma) ** (current_epoch / total_epochs)
\textit{Note:} For ablation purposes, we also attempt to use linear interpolation of $\gamma$ given by :
\begin{align}
    \gamma_{curr} = \gamma_{init} + (\gamma_{fin} - \gamma_{init}) * {(z_{curr} / Z)}
\end{align}

\section{Experimental Setup}
\subsection{Dataset}
% We are dedicated to improving the baseline results for the 
We utilize MECCANO dataset, which comprises egocentric human-object interaction videos in an industrial setting. This dataset, recently introduced and enriched with First Person Vision (FPV) information, holds a wide range of practical applications. The dataset includes Gaze signals, Depth maps, and RGB videos. It offers 20 sequences with 61 action classes. For our experiments, we employ RGB and Depth modalities. The dataset is comprised of 299,376 annotated frames, which includes 61 Action classes and 20 Object classes collected from 20 different participants. For our work, we only utilize the action labels with RGB and Depth modality. We utilize the standard train-val-test split of 55\%-10\%-35\% of the total split.

% These include aiding robotics in the production process, monitoring machine utilization, scheduling calibration tasks, providing guidance to operators, and enhancing safety measures in factories through the prediction of worker actions and interactions with objects.

% We aim to enhance the baseline results for the MECCANO dataset, a multimodal dataset that includes egocentric human-object interaction videos in an industrial environment. 
% This newly introduced dataset, enriched with First Person Vision (FPV) information, has diverse practical applications such as robotic assistance in the production process, monitoring machine usage, scheduling calibration tasks, offering guidance to operators, and enhancing safety measures in a factory by predicting worker actions and interactions with objects. 

\subsection{Data-preprocessing}
We resize the frames to a width of $256$, without disturbing the aspect ratio of the original image, followed by a random crop of $224 \times 224$. We build a clip of 16 consecutive frames and apply random cropping consistently across the clip. In the case of shorter sequences, we pad the sequence with the last frame.

\subsection{Training}
We use the Swin3D-B~\cite{videoswin} backbone, which is pre-trained on the Something-Something v2~\cite{mahdisoltani2018effectiveness} dataset. We adopt focal loss~\cite{lin2017focal} and modify it with exponentially decaying modulating factor $\gamma$ for training the classification model. For optimization, AdamW optimizer with a learning rate of $3 \times 10^{-4}$. 
% and a weight decay of 0.05 has been employed. 
Our model converges in just 20 epochs on the MECCANO dataset. 

\subsection{Evaluation Metrics}
Following the standard performance metrics of MECCANO dataset~\cite{ragusa_MECCANO_2023}, we report the {\it{Top-1}} and {\it{Top-5}} classification accuracy as our evaluation metrics. Additionally, to demonstrate the effectiveness of employing our focal loss variant for this task, we present the class-weighted performance measures across classes: 
% class-weighted average class 
{\it{Precision}}, {\it{Recall}} and {\it{F1-score}}. In our case, we calculate the \textit{F1-score} for each class individually and then compute a weighted average based on class prevalence. This weighted \textit{F1-score} accounts for class imbalance, ensuring appropriate importance is given to each class's performance relative to its representation in the data. More details of the weighted F1-score can be found in ~\cite{ragusa_MECCANO_2023}.

% \noindent\underline{\textit{Top-1} Accuracy}, commonly referred to as accuracy, measures a model's effectiveness in accurately assigning instances to their respective classes. It is calculated as the percentage of instances for which the model's top prediction corresponds to the ground-truth class.

% \noindent\underline{\textit{Top-5} Accuracy} is a more lenient evaluation metric that assesses whether the ground-truth class is amongst the top five predictions.

% \noindent\underline{\textit{Precision}} is the ratio of true positives to the sum of true positives and false positives for a specific class. It quantifies the model's ability to identify instances belonging to a particular class, among all the instances that it has predicted as that class.

% \noindent\underline{\textit{Recall}} is the ratio of true positives to the sum of true positives and false negatives for a given class. It measures the capacity of the model to capture all instances of a specific class, among the actual instances of that class. 

% \noindent\underline{\textit{F1-score}} is the harmonic mean of precision and recall. In our case, we calculate the \textit{F1-score} for each class individually and then compute a weighted average based on class prevalence. This weighted \textit{F1-score} accounts for class imbalance, ensuring appropriate importance is given to each class's performance relative to its representation in the data.
 
\subsection{Comparison To State-Of-The-Art}
Table \ref{table:Sota} presents a comprehensive comparison of the performance of various methods on the MECCANO dataset, a benchmark dataset for industrial-like scenarios in multimodal action recognition. Notably, our method stands out with a {\it Top-1} accuracy of 52.82\% and a {\it Top-5} accuracy of 83.85\%, suggesting its superior performance compared to the state-of-the-art methods listed in the table. The ICIAP 2023 multimodal action recognition challenge leaderboard can be found online at \url{https://iplab.dmi.unict.it/MECCANO/challenge.html}.
\setlength{\tabcolsep}{6pt}
\begin{table}[!th]
\begin{center}
\renewcommand{\arraystretch}{1}
\begin{tabularx}{8.5cm}{l*{13}{c}{c}}
\toprule
{\textbf{Team}} & {\textbf{Modality}} & \multicolumn{2}{c}{\textbf{Accuracy}} \\
\cmidrule(lr){3-4}      
& & {\it Top-1} & {\it Top-5} \\
\midrule
UNICT~\cite{ragusa2021meccano} &	RGB+Depth &	49.49 &	77.61 \\
TORONTO~\cite{liu2023gaddccanet}	& - & 49.52 & 74.21 \\
UNICT~\cite{ragusa2021meccano} & RGB+Depth+Gaze & 49.66 & 77.82 \\
MACAU / UMDR ~\cite{zhou2023unified,zhou2023unifiedpami}  & RGB+Depth & 50.30 & 78.46 \\
LUBECK~\cite{hu2023uniformer} & - & 51.82 & \secondbest{83.35} \\
UNIBZ~\cite{bianchi2023gate-shift-fuse} & - & \secondbest{52.57} & 81.53 \\
UCF ({\bf Ours}) & RGB+Depth & \textcolor{red}{\bf 52.82} & \textcolor{red}{\bf 83.85} \\
\bottomrule
\end{tabularx}
\end{center}
\caption{
\textbf{Comparison with state-of-the-art results on the MECCANO dataset.} 
Our work, declared as the \textit{challenge} winner, ranks first on the leaderboard for Multimodal Action Recognition on the MECCANO dataset. The best method is shown in \bestresult{Red} and the second best method is shown in \secondbest{Blue}. \textit{Note: Information about the modalities used in some competing works, marked with `-', is not available at the time of paper submission.}
}
\label{table:Sota}
\vspace{-3mm}
\end{table}
% {\url{https://iplab.dmi.unict.it/MECCANO/challenge.html}} 
% \vspace{15mm}
\setlength{\tabcolsep}{12pt}
% \vspace{15mm}
\begin{table*}[!th]
\begin{center}
\renewcommand{\arraystretch}{1}
\vspace{3mm}
\begin{tabularx}{12.7cm}{l*{13}{c}{c}}
\toprule
{\textbf{Modality}} & {\textbf{Loss}} & \multicolumn{2}{c}{{\textbf{Accuracy}}} & \multicolumn{2}{c}{{\textbf{AVG Class}}} & {\textbf{AVG {\textit{F1-score}}}}  \\
\cmidrule(lr){3-4}                  
\cmidrule(lr){5-6}
& & {\it Top-1} & {\it Top-5} & {\it Precision} & {\it Recall} & \\
\midrule
RGB & {\it{CE}} & {48.35} & 80.91 & 45.52 & 48.35 & 46.22 \\
% Depth & {\it{CE}} & 45.66 & 77.40 & 43.50 & 45.66 & 43.55 \\
Depth & {\it{CE}} & {43.32} & 75.38 & 41.79 & 43.32 & 41.88 \\
% RGB+Depth & {\it{CE}} & 51.61 & 83.28 & 47.98 & 51.61 & 48.30 \\
RGB+Depth & {\it{CE}} & 50.94 & 81.79 & 47.28 & 50.94 & 48.08 \\
\midrule
RGB & Focal & {50.80} & 82.36 & 47.17 & 50.80 & 47.95 \\
Depth & Focal & 45.52 & 78.07 & 43.74 & 45.52 & 43.41 \\
RGB+Depth & Focal & \textcolor{black}{\bf 52.82} & \textcolor{black}{\bf 83.85} & \textcolor{black}{\bf 49.97} & \textcolor{black}{\bf 52.82} & \textcolor{black}{\bf 49.41} \\
\midrule
RGB$^*$ & Focal & 53.03 & 85.37 & 50.46 & 53.03 & 50.39 \\
Depth$^*$ & Focal & 48.39 & 80.55 & 46.43 & 48.39 & 46.35 \\
RGB+Depth$^*$ & Focal & \textcolor{red}{\bf 55.37} & \textcolor{red}{\bf 85.58} & \textcolor{red}{\bf 52.41} & \textcolor{red}{\bf 55.37} & \textcolor{red}{\bf 52.28} \\
\bottomrule
\end{tabularx}
\end{center}
\caption{{\bf Ablation on Cross-Entropy ({\it{CE}}) loss v/s Focal loss with exponentially decaying $\gamma$.} Results demonstrating the effectiveness of our focal loss variant with an exponentially decaying modulating factor for the action recognition task on the MECCANO test dataset. {\bf *} refers to model trained using both train$+$validation set.}
\label{table:Ablation_CEvsFL}
\vspace{-2mm}
\end{table*}
\subsection{Ablation Study}

\begin{figure*}[t]
  \centering
  \vspace{2mm}
  \includegraphics[width=0.92\textwidth]{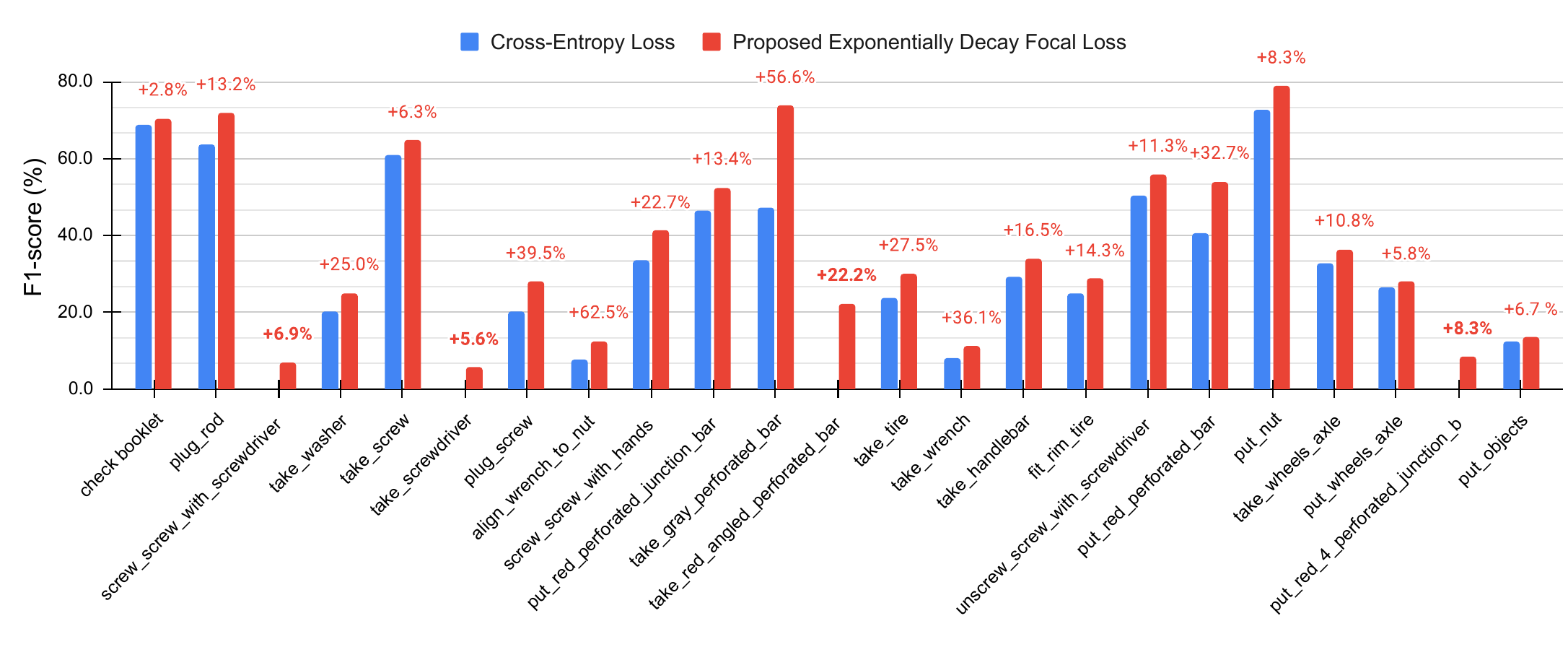}
  \vspace{-4mm}
\caption{\textbf{Class-wise \textit{F1-score} on MECCANO dataset.} We compare class-wise \textit{F1-score} for standard cross-entropy loss vs. our proposed exponentially decaying focal loss. \bestresult{+number} shows the relative improvement of our proposed loss over the cross-entropy loss. Our proposed loss consistently improves the performance of the tail-classes.}
  \label{fig:ClasswiseF1}
  \vspace{-4mm}
\end{figure*}

\noindent\textbf{Cross-Entropy loss v/s Focal loss with exponentially decaying modulating factor:}
Table \ref{table:Ablation_CEvsFL} presents our results on the MECCANO test set. Applying cross-entropy loss to fine-tune our model, pre-trained on Something-Something v2, gives us an initial baseline accuracy of 50.94\% on our multimodal setup. Introducing focal loss with exponential decay in the modulating factor $\gamma$ boosts the overall accuracy by approximately \textbf{2\%} on all performance measures. Furthermore, combining the train and validation data gives the best {\it{Top-1}} accuracy of 55.37\%. \\
Apart from overall improved performance, when we look at the class-wise performance in Fig.~\ref{fig:ClasswiseF1}, our proposed loss function consistently improves the tail class performance, and in some action-classes (\textit{screw\_screw\_with\_screwdriver}, \textit{take\_screwdriver}, and \textit{put\_red\_4\_perforated\_junction\_bar}), where cross-entropy loss was not even able to get any samples of the correct predictions, our loss is able to predict them correctly. 

% Another evidence can be observed through Fig~\ref{fig:ClasswiseF1} of our proposed loss function over standard Cross Entropy loss. Our method improves the recall for each class in the long-tailed distribution of the MECCANO dataset.

\begin{figure}[h]
\begin{center}
   \includegraphics[width=0.9\linewidth]{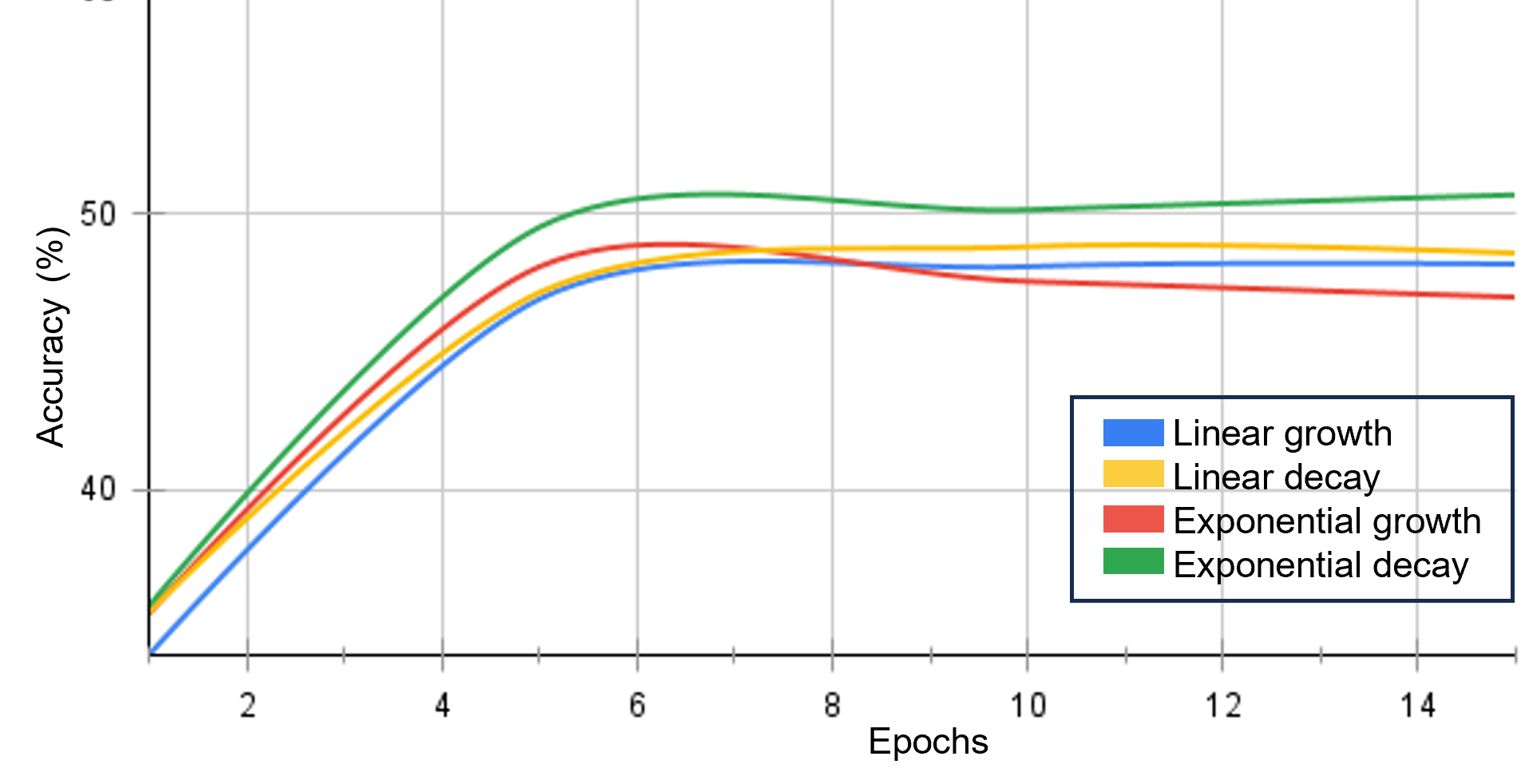}
\end{center}
\vspace{-4mm}
   \caption{{\bf Graphical representation of variations in the focal loss modulating factor $\gamma$.} Illustration depicts the effect of different values of $\gamma$ on the {\it Top-1} action recognition accuracy across the MECCANO test set.}
\label{fig:Ablation_FLVariants}
\vspace{-2mm}
\end{figure}

\noindent\textbf{Design choice of focal loss modulating factor:}
Table \ref{table:Ablation_FLVariants} and Fig.~\ref{fig:Ablation_FLVariants} provide a detailed analysis of the impact of the focal loss modulating factor $\gamma$ on top-1 accuracy in the context of the long-tailed MECCANO dataset. Here, we meticulously examine four distinct profiles of $\gamma$: 

\noindent\textit{\underline{Linear growth}}: linear increase from 0.1 to 2

\noindent\textit{\underline{Linear decay}}: linear decrease from 2 to 0.1 

\noindent\textit{\underline{Exponential growth}}: 
 exponential increase from 0.1 to 2

\noindent\textit{\underline{Exponential decay}}: exponential decrease from 2 to 0.1 

\noindent Each profile represents a unique strategy for changing $\gamma$ with respect to the ongoing number of the training epoch.

\setlength{\tabcolsep}{15pt}
\begin{table}[!th]
\begin{center}
\renewcommand{\arraystretch}{1}
\begin{tabularx}{6cm}{l*{13}{l}{c}}
\toprule
{\bm{{$\gamma$}}} & {\textbf{{\textit{Top-1}} Accuracy}} \\
\midrule
\grayedout{CE baseline} & \grayedout{48.35} \\
\hline
Linear growth &	48.03 \\
Linear decay & 48.46 \\
Exponential growth & 46.76 \\
Exponential decay & 50.80~\increase{(+2\%)} \\
\bottomrule
\end{tabularx}
\end{center}
\caption{{\bf Impact of focal loss modulating factor $\gamma$.} Analysis depicting the significance of the introduced exponentially decaying variant of focal loss on action recognition in the long-tailed MECCANO dataset.} 
\label{table:Ablation_FLVariants}
\vspace{-2mm}
\end{table}

% \begin{figure*}[h]
      
%     \begin{subfigure}[t]{0.49\textwidth}
%         % \vspace{1mm}
%         \centering
%         \includegraphics[width=0.90\textwidth]{ICRA2024_Multimodal/images/Confusion_CE_1.png}
%         % \vspace{-2mm}
%         \caption{{Confusion matrix with Cross-Entropy Loss} }

%     \end{subfigure}
%     \hfill
%     \begin{subfigure}[t]{0.49\textwidth}
%     % \vspace{2mm}
%     \centering
%         \includegraphics[width=0.90\textwidth]{ICRA2024_Multimodal/images/Confusion_FL_1.png}
%         % \vspace{-2mm}
%         \caption{{Confusion matrix with our variant of Focal Loss} }

%     \end{subfigure}
%     \hfill
%     % \vspace{-2mm}
%     \caption{The resultant confusion matrix obtained from the MECCANO test set highlights our model's proficiency in handling the long-tailed distribution.}
%     \label{fig:Confusion}
%     \vspace{-3mm}
% \end{figure*}

  % \caption{\textbf{Class-wise \textit{F1-score}.} Graphical representation of class-wise \textit{F1-score} highlighting our model's proficiency in handling the long-tailed distribution.} 
  
\setlength{\tabcolsep}{15pt}
\begin{table}[!th]
\begin{center}
\renewcommand{\arraystretch}{1}
\begin{tabularx}{8.5cm}{l*{13}{l}{c}}
\toprule
{\textbf{Modality}} & {\textbf{Pre-trained backbone}} & {\textbf{{\textit{Top-1}} Accuracy}} \\
\midrule
RGB	& Freeze & 45.90 \\
RGB	& Fine-tune	& 48.35~\increase{(+2\%)} \\
\hline
Depth &	Freeze & 35.92 \\
Depth & Fine-tune & 43.32~\increase{(+7\%)} \\
\bottomrule
\end{tabularx}
\end{center}
\caption{{\bf Effect of freezing pre-trained weights across different modalities.} Experiments indicating improvements in MECCANO dataset results when fine-tuning the pre-trained weights from Something-Something v2. } 
\label{table:Ablation_BackboneFreeze}
\vspace{-5mm}
\end{table}
The exponential decay strategy, which initially emphasizes hard samples (scarce tail classes) and then transitions to easy ones (dominant classes), showcases the most substantial improvements in action recognition accuracy. Strikingly, this approach leads to the highest {\it Top-1} accuracy of 50.80\% for RGB modality. By starting with hard samples, the model is compelled to learn from the most challenging examples in the dataset. This process can lead to the development of robust and discriminative features that are crucial for accurately classifying difficult instances. Consequently, the model becomes more resilient to challenging cases, which can be particularly useful in imbalanced datasets where the minority class often consists of hard-to-distinguish examples. Once the model has effectively tackled these challenging instances, it gradually shifts its focus towards easier samples, drawing upon the knowledge acquired from handling the more challenging cases. This gradual transition serves as a preventive measure against overfitting to the minority class. 

In contrast, the exponential growth approach, which initially addresses easy samples before shifting to more challenging ones, demonstrates the lowest {\it Top-1} accuracy of 46.76\%. While it focuses on simpler instances for an extended period of time, its ability to effectively handle difficult cases seems to be comparatively restricted. It is also likely that by starting with easy samples, the model prematurely saturates on the majority class, leading to a suboptimal solution that overlooks the minority class. These findings underscore the importance of a well-balanced training strategy for the MECCANO dataset. 

\noindent\textbf{Impact of fine-tuning:}
In this context, we evaluate the impact of freezing or fine-tuning pre-trained weights from  Something-Something v2 on our baseline SWIN3D-B model using the standard CE loss (see Table \ref{table:Ablation_BackboneFreeze}). Our experiments indicate improvements, when fine-tuning the model using cross-entropy loss from pre-trained weights from Something-Something v2. The Depth modality, in particular, exhibits a significant performance boost of \textbf{7\%} upon fine-tuning the pre-trained weights, owing to the domain shift. 
\setlength{\tabcolsep}{3pt}
\begin{table}[!th]
\begin{center}
\renewcommand{\arraystretch}{1}
\begin{tabularx}{8cm}{l*{13}{c}{c}}
\toprule
{\textbf{Modality}} & {\textbf{Model}} & {\textbf{Pre-training data}} & {\textbf{{\textit{Top-1}} Accuracy}} \\
\midrule
RGB	& Swin3D-T & Kinetics-400 & 36.52 \\
RGB	& Swin3D-B & Kinetics-400 & 40.17 \\
RGB	& Swin3D-B & Something-Something v2 & 48.35 \\
\hline
Depth & Swin3D-T & Kinetics-400	& 35.92 \\
Depth & Swin3D-B & Kinetics-400	& 36.63 \\
Depth & Swin3D-B & Something-Something v2 & 43.32 \\
\bottomrule
\end{tabularx}
\end{center}
% \vspace{-2mm}
\caption{{\bf Influence of backbone model size and pre-trained weights.} Introducing a backbone with higher capacity has a positive impact on action recognition accuracy, as does using pre-training weights from the egocentric action recognition dataset, namely Something-Something v2.} 
\label{table:Ablation_DiffBackboneNData}
\end{table}

\noindent\textbf{Backbone model capacity and pre-training data:}
As observed in Table \ref{table:Ablation_DiffBackboneNData}, the usage of the Swin3D-B model consistently outperforms the Swin3D-T model in both RGB and Depth modalities. This implies that a deeper model, characterized by a higher number of learnable parameters, proves to be more effective in capturing the nuances of the data, consequently leading to improved accuracy. Furthermore, the choice of pre-training data significantly influences model performance on the MECCANO dataset. Model pre-trained on Something-Something v2 consistently outperforms its counterpart pre-trained on Kinetics-400, showcasing the importance of domain-specific pre-training. The Something-Something v2 dataset covers hand-object interactions from a first-person view, making it similar to the industry-like actions in the MECCANO dataset. On the other hand, the Kinetics dataset, which focuses on third-person views of human actions, isn't as closely aligned to actions that take place in an industry-like setting.

This finding aligns with the MECCANO dataset's dedicated focus on egocentric activities.

\begin{figure}[h]
\begin{center}
   \includegraphics[width=0.5\linewidth]{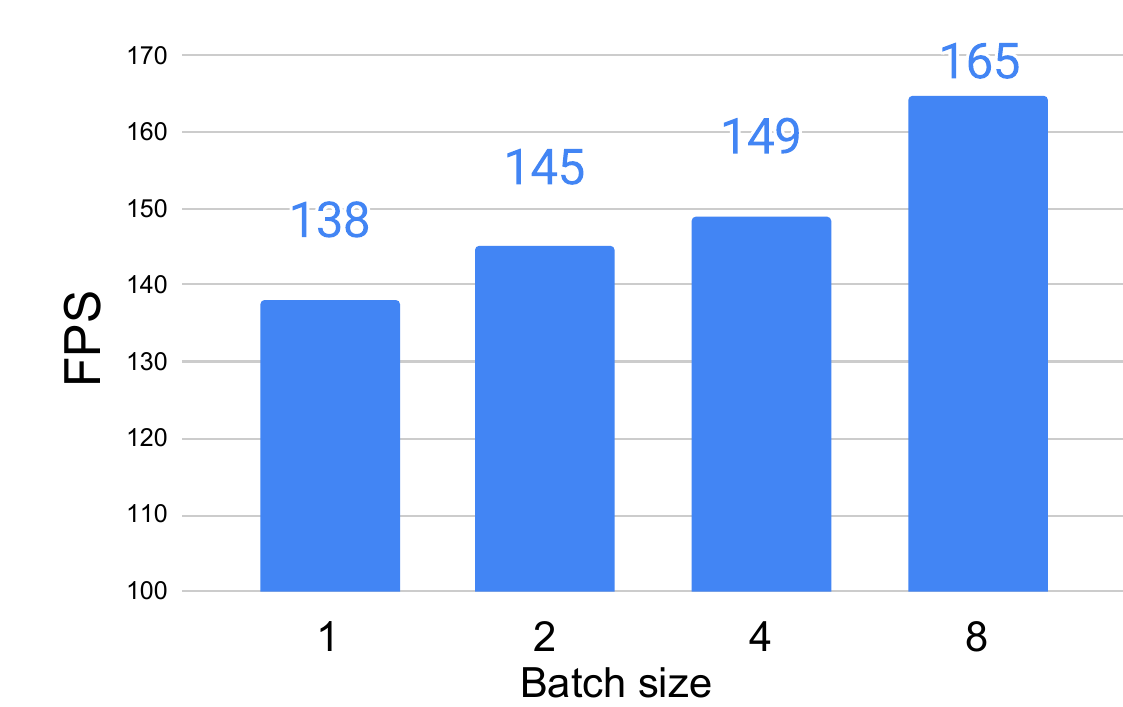}
\end{center}
    \vspace{-3mm}
    \caption{{\bf Runtime.} Frames per second at varying batch-size.}
    \vspace{-5mm}
\label{fig:FPS}
\end{figure}
\noindent\textbf{Runtime computation:}
In order to benchmark the speed of our method, we compute its runtime on a Tesla V100 GPU. We consider both modalities (RGB + Depth) combined for inference and report the frames per second (FPS) in Fig.~\ref{fig:FPS} with varying input batch sizes. Our method attains \textbf{138 FPS} for the combined RGB and Depth modalities, even with a single batch of data, demonstrating its potential for use in real-time systems that typically require 30 FPS.

\section{Conclusion}
In this paper, we proposed a framework to recognize actions from the RGB and Depth multimodal egocentric camera. In order to handle the long-tailed distribution of the action classes, we propose an effective training strategy using the focal loss with an exponentially decaying modulating factor. Our method has set a new state-of-the-art on the MECCANO industry-like dataset and secured first place in the ICIAP 2023 multimodal action recognition challenge. 

We are planning to open-source our code, which can be utilized as a strong baseline for research in video-understanding for multimodal egocentric cameras, especially for industry-like settings. One promising direction for future research lies in integrating the Gaze modality available in datasets like MECCANO, which captures the movement of the human eye while performing the actions. Utilizing Gaze data, alongside hand-object interactions of RGB and Depth modalities, can offer an even deeper understanding of human behavior, providing richer context and enhancing the accuracy on action recognition systems. 
% \begin{figure}
% \begin{center}
%    \includegraphics[width=0.5\linewidth]{Meccano/images/ConfusionMatrix.png}
% \end{center}
%    \caption{The resultant confusion matrix obtained from the MECCANO test set highlights our model's proficiency in handling the long-tailed distribution.}
% \label{fig:Confusion}
% \end{figure}

\clearpage
% \mbox{~}
% \clearpage
% \balance
\bibliographystyle{IEEEtran}
\bibliography{IEEEabrv,egbib}

\end{document}